\documentclass[letterpaper, 10 pt, conference]{ieeeconf}  

\IEEEoverridecommandlockouts                              
\usepackage[autonum,colorhypersetup]{shortex} 
 \definecolor{mpcritic}{RGB}{10, 92, 174}
\usepackage{graphicx}
\graphicspath{{figures/}}

\usepackage{acro}
\acsetup{single={1},	
	uppercase/list}

\DeclareAcronym{AI}{short = AI,
	long = artificial intelligence}
\DeclareAcronym{ML}{short = ML,
	long = machine learning,
	short-indefinite = an}

\DeclareAcronym{SVD}{short = SVD, long = singular value decomposition}
\DeclareAcronym{CCA}{short = CCA,
	long = canonical correlation analysis}	
\DeclareAcronym{FA}{short = FA,
	long = factorial analysis}
\DeclareAcronym{GMM}{short = GMM,
	long = Gaussian mixture model}	
\DeclareAcronym{ICA}{short = ICA,
	long = independent component analysis}	
\DeclareAcronym{LARS}{short = LARS,
	long = least-angle regression}	
\DeclareAcronym{LASSO}{short = LASSO,
	long = least absolute shrinkage and selection operator}	
\DeclareAcronym{LR}{short = LR,
	long = logistic regression}
\DeclareAcronym{PCA}{short = PCA,
	long = principal component analysis}
\DeclareAcronym{PLS}{short = PLS,
	long = partial least squares}	
\DeclareAcronym{RBC}{short = RBC,
	long = reconstruction-based contribution}	
\DeclareAcronym{RMSE}{short = RMSE,
	long = root mean square error}	

\DeclareAcronym{ANFIS}{short = ANFIS,
	long = adaptive network fuzzy inference system}
\DeclareAcronym{ANN}{short = ANN,
	long = artificial neural network,
	short-indefinite = an}	
\DeclareAcronym{BN}{short = BN,
	long = Bayesian network}	
\DeclareAcronym{CNN}{short = CNN,
	long = convolutional neural network}	
\DeclareAcronym{DNNE}{short = DNNE,
	long = decorrelated neural network ensemble}		
\DeclareAcronym{DNN}{short = DNN,
	long = deep neural network}	
\DeclareAcronym{ELM}{short = ELM,
	long = extreme learning machine}
\DeclareAcronym{GAN}{short = GAN,
	long = generative adversarial network}
\DeclareAcronym{GPR}{short = GPR,
	long = Gaussian process regression}	
\DeclareAcronym{GRNN}{short = GRNN,
	long = general regression neural network}	
\DeclareAcronym{MLP}{short = MLP,
	long = multilayer perceptron,
	short-indefinite = an}
\DeclareAcronym{RBFNN}{short = RBFNN,
	long = radial basis function neural network,
	short-indefinite = an}
\DeclareAcronym{RNN}{short = RNN,
	long = recurrent neural network,
	short-indefinite = an}
\DeclareAcronym{RT}{short = RT,
	long = regression tree,
	short-indefinite = an}			
\DeclareAcronym{RVM}{short = RVM,
	long = relevance vector machine,
	short-indefinite = an}		
\DeclareAcronym{SFA}{short = SFA,
	long = slow feature analysis}	
\DeclareAcronym{SVM}{short = SVM,
	long = support vector machine}
\DeclareAcronym{TL}{short = TL,
	long = transfer learning}	
\DeclareAcronym{VAE}{short = VAE,
	long = variational autoencoder}	
\DeclareAcronym{WNN}{short = WNN,
	long = wavelet neural network}				

\DeclareAcronym{RL}{short = RL, long = reinforcement learning, short-indefinite = an}
\DeclareAcronym{A3C}{short = A3C,
	long = asynchronous advantage actor-critic}	
\DeclareAcronym{ADP}{short = ADP,
	long = approximate dynamic programming}
\DeclareAcronym{IQC}{short = IQC, long = integral quadratic constraint}
\DeclareAcronym{BIBO}{short = BIBO, long = {bounded-input, bounded-output}}
\DeclareAcronym{SISO}{short = SISO, long = {single-input, single-output}}
\DeclareAcronym{MIMO}{short = MIMO, long = {multiple-input, multiple-output}}
\DeclareAcronym{DDPG}{short = DDPG,
	long = deep deterministic policy gradient}
\DeclareAcronym{DPG}{short = DPG,
	long = deterministic policy gradient}
\DeclareAcronym{DQN}{short = DQN,
	long = deep $Q$-network}
\DeclareAcronym{HJB}{short = HJB,
	long = {Hamilton-Jacobi-Bellman}}
\DeclareAcronym{MPC}{short = MPC, 
	long = model predictive control,
	long-plural-form = model predictive controllers,
	short-indefinite = an}		
\DeclareAcronym{PI2}{short = $\text{PI}^2$,
	long =policy improvement with path integrals}
\DeclareAcronym{PID}{short = PID, 
	long = proportional-integral-derivative}
\DeclareAcronym{PI}{short = PI, 
	long = proportional-integral}
\DeclareAcronym{PPO}{short = PPO,
	long = proximal policy optimization}
\DeclareAcronym{REINFORCE}{short = REINFORCE,
	long = {\emph{RE}ward Increment $=$ \emph{N}onnegative \emph{F}actor $\times$ \emph{O}ffset \emph{R}einforcement $\times$ \emph{C}haracteristic \emph{E}ligibility}}
\DeclareAcronym{RTO}{short = RTO,
	long = real-time optimization,
	short-indefinite = an}	
\DeclareAcronym{SAC}{short = SAC,
	long = soft actor-critic}	
\DeclareAcronym{TD3}{short = TD3,
	long = twin-delayed DDPG}	
\DeclareAcronym{HER}{short = HER,
	long = hindsight experience replay}

\DeclareAcronym{GP}{short = GP,
	long = Gaussian process,
	long-plural-form = Gaussian processes}
\DeclareAcronym{RBF}{short = RBF,
	long = radial basis function,
	short-indefinite = an}	
\DeclareAcronym{SAE}{short = SAE,
	long = sparse autoencoder}	
\DeclareAcronym{DBN}{short = DBN,
	long = deep belief network}	
\DeclareAcronym{LSTM}{short = LSTM,
	long = long short-term memory}	
\DeclareAcronym{KL}{short = KL,
	long = Kullback-Leibler}
\DeclareAcronym{MDP}{short = MDP,
	long = Markov decision process,
	long-plural-form = Markov decision processes,
	short-indefinite = an}
\DeclareAcronym{LQR}{short = LQR, 
	long = linear quadratic regulator}
\DeclareAcronym{DARE}{short = DARE, 
	long = discrete algebraic Riccati equation}
\DeclareAcronym{LTI}{short = LTI, 
	long = linear time-invariant,
	short-indefinite = an}
\DeclareAcronym{GPS}{short = GPS,
	long = guided policy search}	
\DeclareAcronym{GRU}{short = GRU,
	long = gated recurrent unit}	
\DeclareAcronym{ESN}{short = ESN,
	long = echo state network}	
\DeclareAcronym{ENN}{short = ENN,
	long = Elman neural network}	
	
\DeclareAcronym{CSTR}{short = CSTR,
	long = continuous stirred tank reactor}

\usepackage{algorithm}
\usepackage[noend]{algorithmic}

\usepackage{booktabs}

\crefname{equation}{}{} 
\Crefname{equation}{Equation}{Equations} 
\Crefname{figure}{Fig.}{Figs.} 

\newcommand{\mpcritic}{{\texttt{MPCritic}}\xspace}

\title{\LARGE \bf
MPCritic: A Plug-and-Play MPC Architecture for Reinforcement Learning
}

\author{Nathan P. Lawrence\textsuperscript{\textdagger}, Thomas Banker\textsuperscript{\textdagger}, and Ali Mesbah
\thanks{The authors are with the 
        Department of Chemical and Biomolecular Engineering, 
        University of California, Berkeley, CA 94720, USA.
        {\tt\small mesbah@berkeley.edu}}%
        \thanks{This work was supported by the U.S. Department of Energy, Office of Science, Office of Fusion Energy Sciences under award DE‐SC0024472.}
        \thanks{\textsuperscript{\textdagger}These authors contributed equally to this work.}
}

\begin{document}

\maketitle
\thispagestyle{empty}
\pagestyle{empty}

\begin{abstract}
The reinforcement learning (RL) and model predictive control (MPC) communities have developed vast ecosystems of theoretical approaches and computational tools for solving optimal control problems.
Given their conceptual similarities but differing strengths, there has been increasing interest in synergizing RL and MPC.
However, existing approaches tend to be limited for various reasons, including computational cost of MPC in an RL algorithm and software hurdles towards seamless integration of MPC and RL tools.
These challenges often result in the use of ``simple'' MPC schemes or RL algorithms, neglecting the state-of-the-art in both areas.
This paper presents \mpcritic, a machine learning-friendly architecture that interfaces seamlessly with MPC tools.
\mpcritic utilizes the loss landscape defined by a parameterized MPC problem, focusing on ``soft'' optimization over batched training steps; thereby updating the MPC parameters while avoiding costly minimization and parametric sensitivities.
Since the MPC structure is preserved during training, an MPC agent can be readily used for online deployment, where robust constraint satisfaction is paramount.
We demonstrate the versatility of \mpcritic, in terms of MPC architectures and RL algorithms that it can accommodate, on classic control benchmarks.
\end{abstract}

\section{Introduction}
\acresetall

\Ac{RL} and \ac{MPC} have emerged as two successful frameworks for solving optimal control problems.
Each community has developed a mature theory and set of computational tools for dealing with the well-known intractability of dynamic programming \cite{bertsekas2022lessons, bertsekas_neuro-dynamic_1996, rawlings2017model}.
Given their individual success and roots in dynamic programming, there is growing interest in developing complementary frameworks that can synergize the safe decision-making of \ac{MPC} with the flexible learning of \ac{RL} \cite{reiter2025SynthesisModel,lawrence2025Viewlearning}.

\ac{MPC} takes an optimization-based approach to control wherein predictions are made online to select actions.
This strategy is amenable to theoretical results regarding safe system operation, such as stability and robustness \cite{mayne2000Constrainedmodel}, typically through its interpretable structure and reliance on constrained optimization \cite{rawlings2017model}.
Meanwhile, \ac{RL} is an iterative, sample-based framework in which a control policy is learned through trial and error in an uncertain environment.
Broadly, RL schemes consist of theoretical principles, such as policy gradients and $Q$-learning, combined with general-purpose function approximators \cite{sutton_reinforcement_2018}.

While sophisticated software tools have been developed to address the implementation intricacies of \ac{RL} and \ac{MPC} individually,
two significant hurdles arise when combining them:
the cost of running and differentiating \ac{MPC} in an \ac{RL} algorithm; and interfacing the highly specialized tools of \ac{MPC} and \ac{RL}.
Resolving these obstacles would open the door for leveraging the theoretical properties of \ac{MPC} with the scalability of \ac{RL}.
To this end, we propose \mpcritic: an architecture that integrates seamlessly with machine learning and \ac{MPC} tools, allowing for incorporating \ac{MPC} theory in \ac{RL}.
\mpcritic utilizes the interpretable structure of \ac{MPC}---model, cost, constraints---to define a ``critic'' network, a common object in \ac{RL}, while, crucially, avoiding solving the \ac{MPC} problem during training iterations.
Core to \mpcritic is a ``fictitious'' controller that is cheap to evaluate, enabling batched training like any other critic network in \ac{RL}.
Due to the preserved \ac{MPC} structure, the \ac{MPC} can still be solved in real-time, where online control planning and robust constraint satisfaction can be critical. 

The modularity of \mpcritic allows for a range of configurations wherein individual \ac{MPC} components, such as dynamic model and cost, are designed to ensure theoretical properties of the online \ac{MPC}, or possibly learning all components in unison as a more general \ac{RL} function approximator.
Comparing \mpcritic to standard \ac{MPC} and deep \ac{RL} approaches, two configurations are demonstrated:
learning the theoretically-optimal \ac{MPC} for the \ac{LQR} offline,
extending to the online setting with constraints,
and learning a stochastic ``actor'' parameterized by the fictitious controller embedded within \mpcritic for improved performance and constraint satisfaction in a nonlinear environment. Our contributions are as follows:
\begin{itemize}
    \item An algorithmic framework for integrating \ac{MPC} and \ac{RL} that is agnostic to the \ac{RL} algorithm, yet capable of seamlessly incorporating \ac{MPC} theory.
    \item Detailed account of \mpcritic software and implementation, utilizing advanced \ac{RL} and \ac{MPC} tools.
    \item Case studies demonstrating the theoretical connection, scalability, and flexibility of \mpcritic.
\end{itemize}

\section{Background}

\subsection{Markov decision processes}

We consider an \emph{agent} interacting with a dynamic \emph{environment} with state space $\state$ and action space $\action$.
For any state $s \in \state$, an action $a \in \action$ is selected by the agent, leading to a new state $s' \in \state$.
In particular, we write $s' \sim \pp{p}{s'}{s,a}$, assuming the state transition density $p$ satisfies the \emph{Markov property}.
The desirability of a state-action tuple is characterized by a \emph{reward} function $r: \state \times \action \to \reals$.
Writing $r_t = r(s_t, a_t)$ leads to a trajectory
$\{ s_0, a_0, r_0, s_1, \ldots, s_t, a_t, r_t, s_{t+1}, \ldots \}$.
The utility of a trajectory is characterized by the discounted return of future rewards $
\sum_{t=0}^\infty \gamma^t r(s_t,a_t)$,
where $\gamma \in (0,1)$ is a constant.
The link between states and actions is known as a \emph{policy} $\pi$, a probability density where $a \sim \pp{\pi}{a}{s}$.
The agent implements and adapts the policy $\pi$, aimed at improving its expected returns.
Mathematically, this setup is \iac{MDP} and can be framed as
\begin{equation}
\begin{aligned}
    &\text{maximize} && J(\pi) = \mathbb{E}_{\pi}\left[ \sum_{t=0}^{\infty} \gamma^{t}r (s_t,a_t) \right]\\
    &\text{over all} && \text{policies } \pi \colon \state \to \mathcal{P}(\mathcal{A}),
\end{aligned}
\label{eq:mdpobjective}
\end{equation}
where $\mathcal{P}(\mathcal{A})$ is the set of probability measures on $\action$ and the expectation is over trajectories generated by $\pi$.

In tackling \cref{eq:mdpobjective}, it is useful to define the state-action value function, or $Q$-function, for a policy $\pi$:
	$Q^{\pi} \left( s, a \right) = \mathbb{E}_{\pi}\left[ \sum_{t=0}^{\infty} \gamma^{t} r (s_t, a_t) \middle| s_0 = s, a_0 = a \right]$.
Value functions are an essential ingredient for solving the \ac{MDP} problem in \cref{eq:mdpobjective}.
In particular, they lead to the \emph{Bellman optimality equation}
\begin{equation}
Q^\star (s,a) = r (s,a) + \gamma \EE_{s' \sim \pp{p}{s'}{s,a}} \left[ \max_{a' \in \action} Q^\star (s',a') \right].	
\label{eq:bellmanQ}
\end{equation}
Namely, an optimal policy is designed through ``greedy'' optimization of the optimal value function
\[
\pi^\star (s) = \argmax_{a} Q^\star (s,a).
\label{eq:optimalpi}
\]
Obtaining $Q^\star$ and $\pi^\star$ exactly is generally intractable due to lack of precise knowledge of the transition dynamics and complications surrounding the expectation and maximization operators \cite{bertsekas2012dynamic}.
Nonetheless, \cref{eq:bellmanQ,eq:optimalpi} serve as fundamental inspiration for \ac{RL} and \ac{MPC}.


\subsection{Reinforcement learning}


Although we cannot obtain $Q^\star$ and $\pi^\star$ directly, if we had some oracle mapping $\pi \to Q^\pi$, then an even better policy $\pi^{+}$ could be derived as
$\pi^{+} (s) = \argmax_{a} Q^{\pi} (s,a)$.
This is a general recipe: acquire $Q$, maximize it, and repeat.


In practice, we consider two parameterized function approximators: $Q_\phi$ and $\pi_\theta$, where $\phi$ and $\theta$ are sets of trainable parameters.
The \emph{critic} $Q_\phi$ is trained to satisfy \cref{eq:bellmanQ}; the \emph{actor} $\pi_\theta$ is tasked with both exploring the environment and maximizing $Q_\phi$.
For exploration, $\pi_\theta$ has the form
\[
\pp{\pi}{a}{s} = \mathcal{N} \left( \mu_\theta (s), \Sigma \right),
\label{eq:noisepi}
\]
where the mean is parameterized by the deterministic policy $\mu_\theta$.
Moreover, $\mu_\theta$ is trained such that
\[
Q_\phi (s, \mu_\theta(s)) \approx \max_{a \in \action} Q_\phi (s, a).
\]
The left-hand side is a simple function evaluation, while the right-hand side requires an optimization routine. 
An iterative sequence then follows
\begin{subequations}
\begin{align}
	q &= r + \gamma  Q_\phi (s', \mu_\theta(s'))\label{eq:dpgalg_target}\\
	\phi &\leftarrow \phi - \alpha \grad_\phi \frac{1}{\abs{\mathcal{D}}} \sum_{(s, a, r, s') \in \mathcal{D}} \left( Q_\phi (s,a) - q \right)^2\label{eq:dpgalg_critic}\\
	\theta &\leftarrow \theta + \alpha \grad_\theta \frac{1}{\abs{\mathcal{D}}}  \sum_{(s, a, r, s') \in \mathcal{D}} Q_\phi (s, \mu_\theta (s))\label{eq:dpgalg_actor}.
\end{align}
\label{eq:dpgalg}
\end{subequations}

Equation \eqref{eq:dpgalg_target} is a target sample of the right-hand side of \cref{eq:bellmanQ}.
By collecting a dataset $\mathcal{D}$ of transition tuples $(s, a, r, s')$, the critic weights are updated in \cref{eq:dpgalg_critic} to minimize the residual based on \cref{eq:bellmanQ}.
Finally, the actor is updated in \cref{eq:dpgalg_actor} to improve its maximization performance.
Collectively, \cref{eq:dpgalg} represents the nominal equations comprising the \ac{DPG} algorithm \cite{silver2014DeterministicPolicy}; these ideas then led to deep \ac{RL} algorithms such as \acs{TD3} \cite{fujimoto2018addressing} and \acs{SAC} \cite{haarnoja2018Softactorcritic}.

\subsection{Model predictive control parameterization}

\Ac{MPC} takes a different view towards solving \acp{MDP}.
At each time step, it uses a dynamic model of the environment, cost, and constraints to plan a sequence of actions.
The first action is applied to the environment, and the process is repeated; this is a \emph{receding horizon} approach to control.

\Ac{MPC} considers the following value parameterization
\begin{equation}
\begin{aligned}
   Q_\phi (s,a) =  &\underset{u_0, \ldots, u_{N-1}}{\text{min}} && \sum_{t=0}^{N-1} \ell(x_t, u_t) + V(x_N) \\
    &\text{subject to} && x_0 = s,\quad u_0 = a \\
    & && x_{t+1} = f(x_t, u_t) \\
    & && h(x_t,u_t) \leq 0,\quad g(u_t) \leq 0.
\end{aligned}
\label{eq:Qmpc}
\end{equation}
This leads to the deterministic policy $\mu (s) = \argmin_{a} Q_\phi (s,a)$.
$\phi$ encompasses parameters of the dynamic model $f$, stage cost $\ell$, and terminal value function $V$.
\Cref{eq:Qmpc} is a modular structure, meaning individual components may be fixed or modified by different means. 
For instance, the dynamic model may be derived from system identification.
Each element of $\phi$ is designed such that \cref{eq:Qmpc} is a tractable approximation of the \ac{MDP} problem.

\Ac{MPC} provides an interpretable representation for approximating $Q^\star$.
Its model-based structure enables safety and robustness properties \cite{rawlings2017model}, making it a desirable parameterization for learning-based control.

\section{Algorithmic framework of \mpcritic}

\subsection{MPC-based architecture via fictitious controller}

\begin{figure}[tbp]
\centering
\includegraphics[scale=1.0]{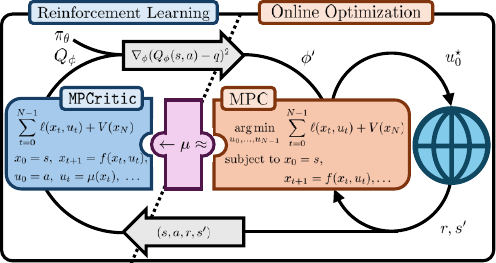}
\caption{
\mpcritic ``plugs'' into \ac{RL} and \ac{MPC} tools.
Left: The fictitious controller $\mu$ approximates the \ac{MPC} optimization in \mpcritic for efficient RL with critic $Q_{\phi}$ and actor $\pi_{\theta}$ for generating targets $q$.
Right: The modified critic parameters $\phi'$ are transferred to the exact, online \ac{MPC} formulation to then gather transition tuples $(s,a,r,s')$ for further refining $\phi$.
}
\label{fig:concept}
\end{figure}

The proposed architecture, \mpcritic, utilizes a so-called ``fictitious'' controller to take the space of the decision variables $\{ u_1, \ldots, u_{N-1}\}$ in the \ac{MPC} problem in \cref{eq:Qmpc}.
Over some restricted domain of the state-action space, consider the following $Q$-function parameterization
\begin{equation}
\begin{aligned}
    Q_\phi (s,a) &= \frac{1}{N} \sum_{t=0}^{N-1} \ell(x_t, u_t) + V(x_N) \\
    \text{Given} &
    \begin{cases} 
    \begin{aligned}
    & x_{t+1} = f(x_t, u_t) && x_0 = s\\
    & \textcolor{mpcritic}{ u_t = \mu (x_t)} && u_0 = a \\
    &  h(x_t,u_t) \leq 0,\quad g(u_t) \leq 0,
    \end{aligned}
    \end{cases}
\end{aligned}
\label{eq:mpcritic_nominal}
\end{equation}
where $\phi = \{ \ell, V, f, \mu\}$.
The controller $\mu$ is fictitious because it never interacts with the environment.
Instead, it serves several important functions:
\begin{enumerate}
	\item \textbf{Efficiency.}\quad Querying $Q_\phi$ only requires running the system model forward and accumulating the closed-loop cost.
	\item \textbf{Approximation.}\quad $\mu$ is trained to approximate the minimization step that \ac{MPC} requires.
	\item \textbf{Modularity.}\quad $Q_\phi$ contains the \ac{MPC} structure and can be seamlessly integrated with \ac{RL} tools. Yet, at deployment, $\mu$ is disregarded and the full \ac{MPC} optimization is performed online. 
\end{enumerate}
Taken together, $\mu$ enables batched, iterative training in \iac{RL} ecosystem, while preserving the exact \ac{MPC} structure for online control, as shown in \cref{fig:concept}.

\Cref{eq:mpcritic_nominal} is not defined over the full state-action space.
Therefore, in practice, \mpcritic takes the form of
\begin{equation}
\begin{multlined}
	Q_\phi (s,a) = \frac{1}{N} \Bigg( \sum_{t=0}^{N-1} \ell(x_t, u_t) + V(x_N)  \\ + \rho \sum_{t=0}^{N-1} \norm{ \max\left\{ h(x_t, u_t), 0 \right\} } \Bigg),
\end{multlined}
\label{eq:mpcritic}
\end{equation}
where $\rho > 0$ is a constant penalty term. The system dynamics and fictitious controller are explicitly accounted for in computing \cref{eq:mpcritic}; the action constraints are part of the architecture $\mu$.
A penalty approach is assumed in \cref{eq:mpcritic} for simplicity.
However, the proposed setup is general, inviting other approaches such as barrier or augmented Lagrangian methods.

\textbf{Relation to differentiable MPC.}\quad
An alternative to \mpcritic is to directly differentiate through the \ac{MPC} solution \cite{amos2019DifferentiableMPC,gros2020DataDrivenEconomica}.
This line of work has the benefit of preserving the \ac{MPC} structure, while modifying it under some supervisory signal, such as reward or an imitation loss. 
However, embedding the \ac{MPC} optimization routine into a general \ac{RL} framework is cumbersome and expensive because the number of \ac{MPC} solves scales with the number of time steps, update iterations, and batch size.
Our approach treats \ac{MPC} as a loss, allowing for approximate solutions driving batched parameter updates, but preserves the \ac{MPC} structure for online deployment.

\textbf{Relation to approximate MPC.}\quad
The so-called fictitious controller in \cref{eq:mpcritic} aims to approximate the minimization process of \ac{MPC}.
This is conceptually similar to approximate \ac{MPC} \cite{chen2018ApproximatingExplicit,karg2020EfficientRepresentation,paulson2020approximate,drgona2024LearningConstrained}.
In fact, the proposed approach can be viewed as a combination of approximate \ac{MPC} and actor-critic methods in \ac{RL} \cite{konda1999ActorcriticAlgorithms}.
Instead of using $\mu$ to decrease online computational demand, we use it to integrate the \ac{MPC} structure into \ac{RL}.
Consequently, $\mu$ is never trained for high accuracy over the state-action space for a particular \ac{MPC} configuration $\phi$.
Rather, it is part of a dynamic cycle of refinements to $\phi$ and $\theta$, as in actor-critic methods.

\subsection{\mpcritic learning configurations}
\label{subsec:configs}

Different variants of \mpcritic depend on two factors: 
\begin{enumerate}
    \item \textbf{Role of the fictitious controller and model in \mpcritic.}\quad $\mu$ in \mpcritic may be viewed either as an approximation to the \ac{MPC} solution, or as any other parameter, trained entirely from the reward signal. The same distinction applies to the dynamic model $f$.
    \item \textbf{Definition of the policy.}\quad \mpcritic preserves the online \ac{MPC} agent for control simply by removing $\mu$. Alternatively, \mpcritic may be used solely as a critic network, leaving the opportunity to train a separate actor network for control.
\end{enumerate}
Thus, there are two extreme versions of \mpcritic.
The one presented so far trains $\mu$ to minimize the loss defined by the \ac{MPC} objective, acquires $f$ from system identification, and deploys an online \ac{MPC} policy derived from \mpcritic.
The other extreme trains $\mu$ and $f$ entirely from reward, like arbitrary parameters in a critic network, while training a separate actor network for control.
That is, \mpcritic can, in principle, learn a control-oriented model directly from reward, rather than a system identification method.
The first view is useful when a predefined \ac{MPC} structure is known to be feasible for online control and possibly benefits from favorable theoretical properties, but requires tuning.
The second view does not invoke \iac{MPC} agent and, therefore, does not require an NLP solver, meaning more complex structures may be used in \mpcritic to guide the learning of an easy-to-evaluate actor network; this could be viewed as an adaptive, reward-driven view of approximate \ac{MPC}.

These different configurations are summarized in \cref{alg:RL,alg:mpcritic}. 
$\theta$ and $\phi$ refer to actor and critic parameters, respectively.
Additionally, in light of these different flavors of \mpcritic, we define $\psi$ to be parameters inside \mpcritic that are trained under some auxiliary objective.
Under one view of \mpcritic, we have $\psi = \{ \psi^{(\mu)}, \psi^{(f)} \}$ for the parameters of $\mu$ and $f$ trained in an approximate \ac{MPC} fashion and system identification, respectively.
We may also have $\psi = \emptyset$, meaning we write $\phi^{(\mu)}, \phi^{(f)}$ because $\mu$ and $f$ are part of the set of critic parameters $\phi$.

 \begin{algorithm}[tb]
 \caption{Vanilla RL with \mpcritic}
 \begin{algorithmic}[1]
 \renewcommand{\algorithmicrequire}{\textbf{Input:}}
 \renewcommand{\algorithmicensure}{\textbf{Output:}}
 \STATE Initialize $\theta, \phi, \textcolor{mpcritic}{\psi}$
  \FOR {each environment step}
  \STATE $a \sim \pp{\pi}{a}{s} \hfill\triangleright\ \text{Optional: See \cref{alg:mpcritic}}$
  \STATE $s', r \sim \pp{p}{s', r}{s,a}$
  \FOR {each update step}
  \STATE $\phi \leftarrow \phi - \alpha \nabla \mathcal{L}_{\text{critic}} \hfill\triangleright\ \text{e.g., \cref{eq:dpgalg_critic}}$
  \STATE $\theta \leftarrow \theta + \alpha \nabla \mathcal{L}_{\text{actor}} \hfill\triangleright\ \text{e.g., \cref{eq:dpgalg_actor}}$
  \IF {\textcolor{mpcritic}{$\psi \neq \emptyset$}}
  \STATE $\psi^{(f)} \leftarrow \psi^{(f)} - \alpha \nabla \mathcal{L}_{\text{model}} \hfill\triangleright\ \text{e.g., MSE}$\label{algeq:ID}
  \STATE $\psi^{(\mu)} \leftarrow \psi^{(\mu)} - \alpha \nabla \mathcal{L}_{\text{control}}  \hfill\triangleright\ \text{e.g., \cref{eq:mpcritic}}$\label{algeq:mu}
  \ENDIF
  \ENDFOR 
  \ENDFOR
 \end{algorithmic} 
 \label{alg:RL}
 \end{algorithm}

 \begin{algorithm}[tb]
 \caption{Optimization-based \mpcritic actor}
 \begin{algorithmic}[1]
 \renewcommand{\algorithmicrequire}{\textbf{Input:}}
 \renewcommand{\algorithmicensure}{\textbf{Output:}}
\STATE Current \mpcritic parameters $\phi, \textcolor{mpcritic}{\psi}$
  \FOR {each environment step}
  \STATE $\textcolor{mpcritic}{a = \underset{a}{\argmin}\ Q_{\text{\mpcritic}}(s,a)} \hfill\triangleright\ \text{e.g., \cref{eq:Qmpc}} $
  \STATE $s', r \sim \pp{p}{s', r}{s,a}$
  \STATE Update $\phi, \textcolor{mpcritic}{\psi} \text{ via  \cref{alg:RL}}$
  \ENDFOR
 \end{algorithmic} 
 \label{alg:mpcritic}
 \end{algorithm}

\section{Interfacing deep RL with MPC theory}

\mpcritic permits theoretical properties through its \ac{MPC} structure.
We outline how \mpcritic can leverage general \ac{MPC} formulations within \iac{RL} ecosystem, touching on the theoretical and implementation aspects at play.

\subsection{Robustness and stability}

\mpcritic preserves the \ac{MPC} structure, which makes it amenable to existing theory.
We point to several such avenues that future work should more rigorously investigate.
The importance of a terminal value function for stability and constraint satisfaction is well-established \cite{mayne2000Constrainedmodel,rawlings2017model}.
It is straightforward to incorporate quadratic functions, or Lyapunov neural networks, as a terminal cost in the design of \mpcritic, as is done here.
As such, one may invoke \ac{LQR} or certainty equivalence arguments to construct a stable-by-design architecture \cite{lawrence2020AlmostSurely}.
Robustness is another important aspect of \ac{MPC} safety.
Although \mpcritic is trained on the system of interest, robustness is still important for improved constraint satisfaction, especially in the early stages of training, or if training is halted.
Any robust \ac{MPC} \cite{bemporad2007robust} or stochastic \ac{MPC} \cite{mesbah2016stochastic} method is, in principle, compatible with the \mpcritic framework.

\subsection{Deep RL implementation}

\mpcritic is implemented in NeuroMANCER \cite{Neuromancer2023}, a differentiable programming library for solving optimal control problems.
Because NeuroMANCER is based on PyTorch, \mpcritic interfaces nicely with \ac{RL} packages for training its components.
We use CleanRL \cite{huang2022CleanRLHighquality} since its single-file implementations of \ac{RL} algorithms facilitate transparency.
For deployment of \mpcritic, we use do-mpc \cite{fiedler2023DompcFAIR}, a Python toolbox for \ac{MPC} built around CasADi \cite{Andersson2019}.
Finally, L4CasADi \cite{salzmann2024LearningCasADi} serves as a bridge between PyTorch and CasADi, making it a convenient tool for deploying \iac{MPC} agent with the learned \mpcritic models.
Importantly, \mpcritic is not restricted to any particular \ac{MPC} implementation, solver, or \ac{RL} library.
These toolboxes encapsulate general \ac{MPC} formulations, designed to function as any critic network under the \mpcritic framework.

\section{Case studies}

To demonstrate the theoretical properties of \mpcritic, we investigate the convergence of its learned solutions to the analytical \ac{LQR} solutions.
We then test its computational efficiency for increasingly high-dimensional systems as compared to differentiable \ac{MPC}.
Afterwards, the proposed learning-based control framework is evaluated on two control tasks.
In the first, we evaluate \cref{alg:mpcritic} and the learned fictitious controller, comparing to a standard deep RL agent.
The second demonstrates the flexibility of \mpcritic as a function approximator in \cref{alg:RL}, learning a stochastic decision-making actor in a nonlinear environment with constraints.
All experiments were run on an Apple M3 Pro 11 Core laptop. Codes are available at \url{https://github.com/tbanker/MPCritic}.


\subsection{Offline validation \& scalability of \mpcritic}

    We first study \mpcritic in the context of \ac{LQR}.
    Consider an open-loop unstable linear system $s' = As + Bu$, with quadratic reward $r(s,a) = -s^{\top}Ms - a^{\top}Ra$ (see unstable Laplacian dynamics in \cite{recht2019tour}).
    We assume $M$ and $R$ are known, but the parameters of the model $A,B$, terminal cost $P$, and gain $K$ are uncertain.
    Updating as in \cref{alg:RL}, \mpcritic aims to learn the true, optimal parameters for the system model $A^{\star}, B^{\star}$, terminal cost $P^{\star}$ (from the discrete algebraic Riccati equation), and the corresponding optimal gain $K^{\star}$, using $\mu(x)=-Kx$.
    Updates repeatedly follow \cref{eq:dpgalg_critic} for $\phi = P$, \cref{eq:dpgalg_actor} for $\psi^{(\mu)} = K$, and
    \begin{equation}
        \psi^{(f)} \leftarrow \psi^{(f)} - \alpha \grad_{\psi^{(f)}} \frac{1}{\abs{\mathcal{D}}} \sum_{(s, a, s') \in \mathcal{D}} \left( (As + Ba) - s' \right)^2,\label{eq:valalg_model}
    \end{equation}
    for $\psi^{(f)}=\{ A,B \}$.
    All uncertain parameters are initialized following $\psi^{(f)}, \psi^{(\mu)}, \phi, \sim \mathcal{N}(0,1)$ and learned from $10^5$ transitions $(s,a,r,s')$ following $s, a \sim \mathcal{U}(-1,1)$.

    \begin{figure}[t!]
     \centering
     \includegraphics[scale=1.0]{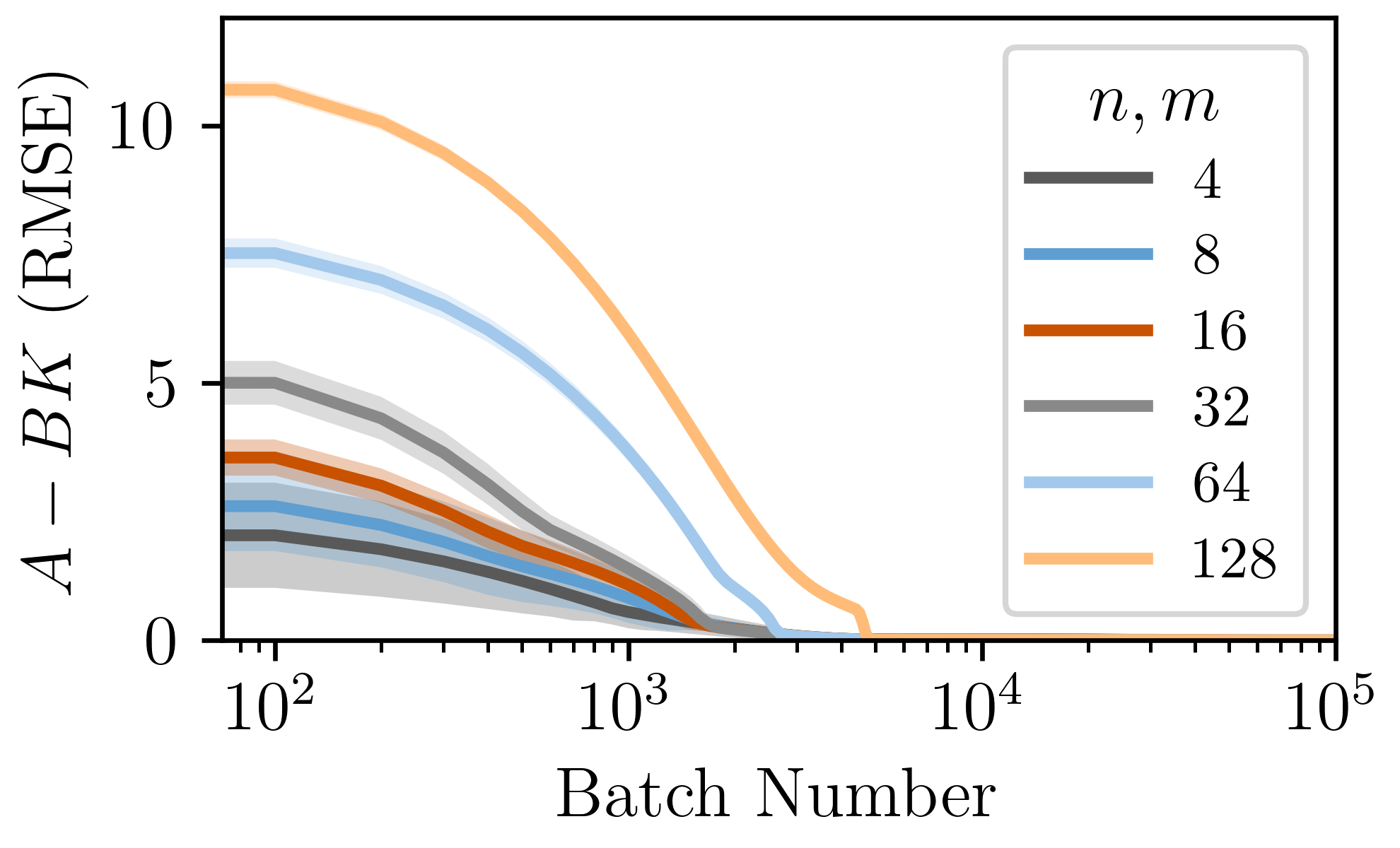}
     \caption{Batched learning of the terminal cost, fictitious controller $K$, and model $A,B$ parameters for equal and increasing state and action dimension, i.e., $n=m$.
     Lines represent the \ac{RMSE} of the learned closed-loop dynamics $A-BK$ with respect to that of the optimal parameter values with $\pm2$ sample standard deviations shaded for $20$ seeds.
     }
     \label{fig:scaling_validation}
    \end{figure}

    The true, optimal parameters are approximately recovered within \mpcritic in this learning scheme for systems of equal and increasing state and action dimension, $n$ and $m$, respectively.
    This is shown in \cref{fig:scaling_validation} in terms of the \ac{RMSE} of the learned closed-loop dynamics $A-BK$ for all systems with each batched update. 
    On average, \ac{RMSE} diminishes to less than $5{\times}10^{-4}$ within $10^{5}$ steps for all system sizes, and although not depicted, that of the model and fictitious controller individually diminish to less than $4{\times}10^{-4}$. 
    Obtaining an accurate representation for the closed-loop dynamics improves \cref{eq:mpcritic} as a $Q$-function approximator, which is further refined by \cref{eq:dpgalg_critic}.
    Accordingly, $P$ is learned such that \cref{eq:mpcritic_nominal} best approximates $Q^{\star}$, resulting in an \ac{RMSE} with respect to $P^{\star}$ less than $2{\times}10^{-2}$ for all systems with the error increasing with system size. 
    This example demonstrates the ability of \mpcritic to (approximately) learn the theoretically-optimal \ac{MPC} components in a batched learning scheme, while preserving the \ac{MPC} structure.

    Additionally, a key benefit of \mpcritic is the computational efficiency of batch processing brought forth by parameterizing the \ac{MPC} optimization through the fictitious controller $\mu$.
    To demonstrate this point, \cref{tab:time_table} reports the average time to evaluate $\mu$, or solve an MPC policy $\pi^{\text{MPC}}$ (forward) and differentiate their outputs (backward).
    For simplicity, the MPC policy is of constrained linear quadratic formulation with horizon $N=1$ and, accordingly, the fictitious controller is defined by a ReLU \ac{DNN} for its piecewise affine structure with $2$ hidden layers of $100$ nodes.
    \Cref{tab:time_table} reports significantly less computation time for the ``soft'' optimization performed by $\mu$ in comparison with the exact optimization of $\pi^{\text{MPC}}$.
    Critically, the backward computation times for $\mu$ are less sensitive to the system size, requiring less than $1$ millisecond in all cases, as compared to $\pi^{\text{MPC}}$ that can take hours to solve and differentiate.
    \Ac{RL} algorithms typically require rapid evaluation of both forward and backward operations, potentially online, quickly making the exact optimization of $\pi^{\text{MPC}}$ and subsequent differentiation impractical for larger systems.
    Rather, cheap evaluation and differentiation, as well as favorable scaling, all while retaining the desired structure of \ac{MPC} through $\mu$, lessen the constraints of computational cost on the user's choice of \ac{RL} algorithm. 

    \begin{table}[btp]
        \centering
        \begin{tabular}{cllll}
            \toprule
            \multicolumn{1}{c}{} & \multicolumn{2}{c}{Forward ($s$)} & \multicolumn{2}{c}{Backward ($s$)} \\
            \cmidrule(rl){2-3} \cmidrule(l){4-5}
                $n,m$ & \quad~$\mu$ & \quad$\pi^{\text{MPC}}$ & \quad~$\mu$ & \quad$\pi^{\text{MPC}}$ \\
            \midrule
                $4$ & $2.4{\times}10^{-4}$ & $4.4{\times}10^{-1}$ & $1.9{\times}10^{-4}$ & $3.1{\times}10^{-1}$\\
                $8$ & $3.0{\times}10^{-4}$ & $5.9{\times}10^{-1}$ & $2.7{\times}10^{-4}$ & $1.4{\times}10^{0}$\\
                $16$ & $3.0{\times}10^{-4}$ & $7.9{\times}10^{-1}$ & $2.1{\times}10^{-4}$ & $9.5{\times}10^{0}$\\
                $32$ & $2.1{\times}10^{-4}$ & $1.5{\times}10^{0}$ & $1.7{\times}10^{-4}$ & $1.3{\times}10^{2}$\\
                $64$ & $2.4{\times}10^{-4}$ & $4.8{\times}10^{0}$ & $1.8{\times}10^{-4}$ & $8.8{\times}10^{2}$\\
                $128$ & $1.1{\times}10^{-3}$ & $1.5{\times}10^{1}$ & $6.0{\times}10^{-3}$ & $9.0{\times}10^{3}$\\
            \bottomrule
        \end{tabular}
        \caption{
        Average time across $10$ seeds to evaluate the forward and backward passes of \iac{DNN} $\mu$ and \iac{MPC} $\pi^{\text{MPC}}$ for a batch of $256$ states, with $s \sim \mathcal{U}(-1,1)$, on systems of equal and increasing state and action dimension, i.e., $n=m$.
        }
        \label{tab:time_table}
    \end{table}

\subsection{Learning \ac{MPC} online with \mpcritic}

    We now explore the application of \mpcritic for learning \ac{MPC} via online interaction and compare it to a traditional deep \ac{RL} agent, both utilizing the TD3 algorithm \cite{fujimoto2018addressing}.
    Consider the previous \ac{LQR} environment with $n=m=4$, initial state $s_{0} \sim \mathcal{U}(-1,1)$, and, now, the goal of maximizing cumulative rewards over an episode of $50$ time steps.
    The \ac{MPC} agent acts by solving a constrained linear quadratic optimization problem, subject to constraints $\norm{x_{t}}_{\infty}\leq1$ and $\norm{u_{t}}_{\infty}\leq1$, with prediction horizon $N=10$.
    All of $\ell$,$V$,$f$, and $\mu$ are learned via \cref{alg:mpcritic}, with auxiliary objectives \cref{eq:mpcritic} and \cref{eq:valalg_model} for $\mu$ and $f$, respectively; $\mu$ being the previously defined ReLU \ac{DNN}. 
    The \ac{RL} agent acts though a ReLU \ac{DNN} policy of the same model class as $\mu$, but is trained to maximize the critic, a neural network with 1 hidden layer of 256 nodes.

    \begin{table}[tbp]
        \centering
        \begin{tabular}{ccccc}
            \toprule
            \multicolumn{1}{c}{} & \multicolumn{2}{c}{Cumulative Reward} & \multicolumn{2}{c}{Constraint Violations} \\
            \cmidrule(rl){2-3} \cmidrule(l){4-5}
                Agent & Mean & SD & Min & Max \\
            \midrule
                \mpcritic & $-64.74$ & $13.87$ & $0$ & $3$\\
                Deep RL & $-135.85$ & $147.38$ & $0$ & $49$\\
            \bottomrule
        \end{tabular}
        \caption{
        Cumulative reward and state constraint violation count statistics of the learned \ac{MPC} and deep \ac{RL} agents during the final $10$ episodes of training across $10$ seeds.
        }
        \label{tab:reward_table}
    \end{table}
    
    Learning for $5{\times}10^{5}$ steps, cumulative reward and constraint violation statistics for both agents during the final $10$ episodes are reported in \cref{tab:reward_table}.
    Notably, the \ac{MPC} agent generally obtains greater rewards with significantly less variance.
    Furthermore, while not shown, the \ac{MPC} agent achieves equal performance, on average, as the final \ac{RL} agent in less than $10^{3}$ update steps.
    This difference in sample efficiency can be attributed, in part, to the auxiliary system identification objective.
    While the deep \ac{RL} agent relies on rewards and bootstrapping to learn the $Q$-function, \eqref{eq:valalg_model} provides an additional complementary signal for improving \mpcritic. 
    \Cref{tab:reward_table} also shows the learned \ac{MPC} agent, in the worst of cases, is less apt to violate $\norm{s_{t}}_{\infty}\leq1$ compared to the deep \ac{RL} agent.
    Although one can attempt to promote this behavior in the \ac{RL} agent by modifying its reward signal, doing so does not readily provide guarantees.
    Rather, \mpcritic provides a straightforward pathway to incorporate state constraints through its architecture.

\begin{figure}[t!]
     \centering
     \includegraphics[scale=0.95]{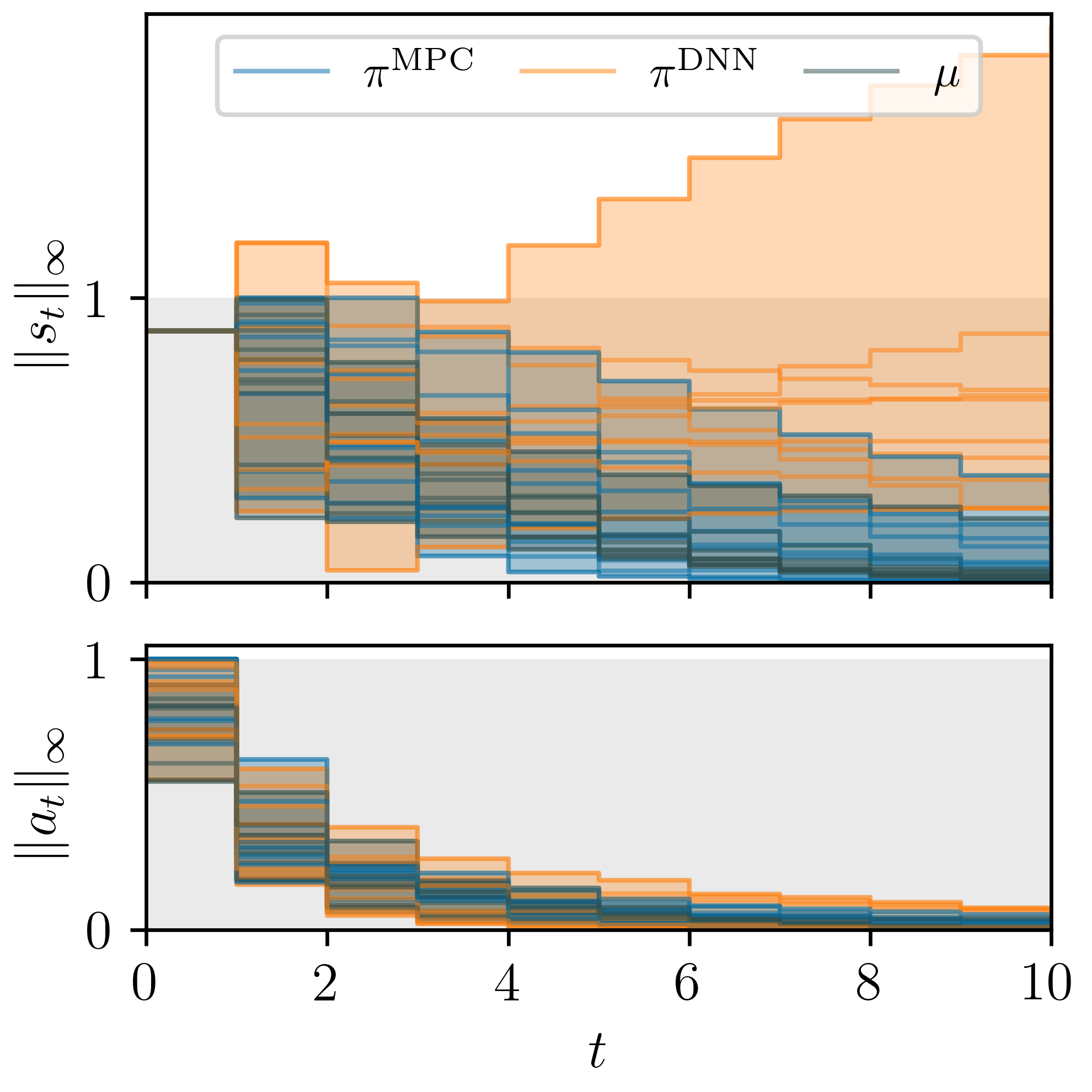}
     \caption{
     Closed-loop trajectories for three control policy classes: a \ac{DNN} actor $\pi^{\text{DNN}}$, the \ac{MPC} of \mpcritic $\pi^{\text{MPC}}$, and $\mu$.
     Ten policies of each class are learned and rolled out from the same initial state, depicting the infinity norm of states $s_t$ and actions $a_t$ with colored shading between policy extrema at each discrete time step $t$.
     All control policies respect the action constraint (gray) by design, but the deep \ac{RL} agent does not readily accommodate the state constraint (gray).
     }
     \label{fig:lqr_traj}
  \end{figure}

    The learned behavior of each policy is shown in \cref{fig:lqr_traj}, including that of the learned fictitious controller $\mu$ for further comparison.
    With rewards penalizing non-zero actions more than states, the RL agent is largely concerned with avoiding large actions rather than driving the state to the origin.
    The \ac{MPC} agent designs coordinated action sequences towards the origin, traversing the boundary of the state and/or action constraints for portions of the sequence.
    In contrast, without a ``planning'' mechanism or modified reward, the deep \ac{RL} agent is willing to leave the closed unit ball to maximize the reward, but this is sure to incur future costs for the unstable system.
    Notably, the \ac{RL} agent's policy and $\mu$ are of the same model class, yet $\mu$ is learned to approximate the exact \ac{MPC} optimization rather than maximize a \ac{DNN} critic.
    Consequently, $\mu$ is informed by the constraints without modifying the reward signal, unlike the \ac{RL} agent, due to their presence in \cref{eq:mpcritic}. 
    This property, along with the relative efficiency of $\mu$, raises interesting questions about how $\mu$ can be leveraged more broadly within \ac{RL} schemes.

\subsection{Maximum entropy policies with \mpcritic}

This example illustrates the generality of \mpcritic as an inductive bias in \ac{RL}.
We use \mpcritic as a function approximator in maximum entropy \ac{RL} \cite{levine2018ReinforcementLearning}.
Here, the goal is to learn a stochastic actor that maximizes its reward, while doing so as randomly as possible.
This randomness induces the exploration useful towards system identification.

Consider a stochastic actor $\pi_\theta$ as in \cref{eq:noisepi} and critic $Q_\phi$, both given by a \ac{DNN}.
\mpcritic is constructed with \iac{DNN} dynamic model that is learned online through system identification, a fixed stage cost, a penalty term for state constraints, and $Q_\phi$ as terminal value function.
In this example, we use neural networks that would be intractable to train using typical NLP solvers.
Instead, the fictitious controller, aimed at minimizing the \ac{MPC} objective in \cref{eq:Qmpc}, parameterizes the mean of $\pi_\theta$.
One can simply run actor-critic update steps on  $\pi_\theta$ and $Q_\phi$, while periodically applying updates to $f$ and $\mu$ as in lines \ref{algeq:ID} and \ref{algeq:mu} of \cref{alg:RL}.
In this way, the policy $\pi_\theta$ learns from the structure of \mpcritic, and \mpcritic adapts with $Q_\phi$.

We demonstrate this learning scheme with \iac{SAC} agent \cite{haarnoja2018Softactorcritic}, an off-policy maximum entropy deep \ac{RL} algorithm.
The environment is modeled with \iac{CSTR}, a common benchmark in process control, and a Gaussian-shaped reward (see \cite{lawrence2025Viewlearning} for further details of this environment).
The goal is to control the concentration $c_B$ to a desired level $c_B^{\text{goal}}$, comprising the reward $r(s,a) = \exp{\left( -\nicefrac{\left( c_B^{\text{goal}} - c_B \right)^2}{2 \sigma^2}\right)}$ with $\sigma^2 = 0.0025$.
This reward structure is used for the stage cost with $\sigma^2 = 0.25$; all other models involved---for $\pi_\theta$, $Q_\phi$, $f_\psi$---are two-layer ReLU networks with $256$ nodes per layer; note $\mu_\psi = \mu_\theta$, which parameterizes the mean of $\pi_\theta$.

\begin{figure}[t!]
     \centering
     \includegraphics[scale=1.0]{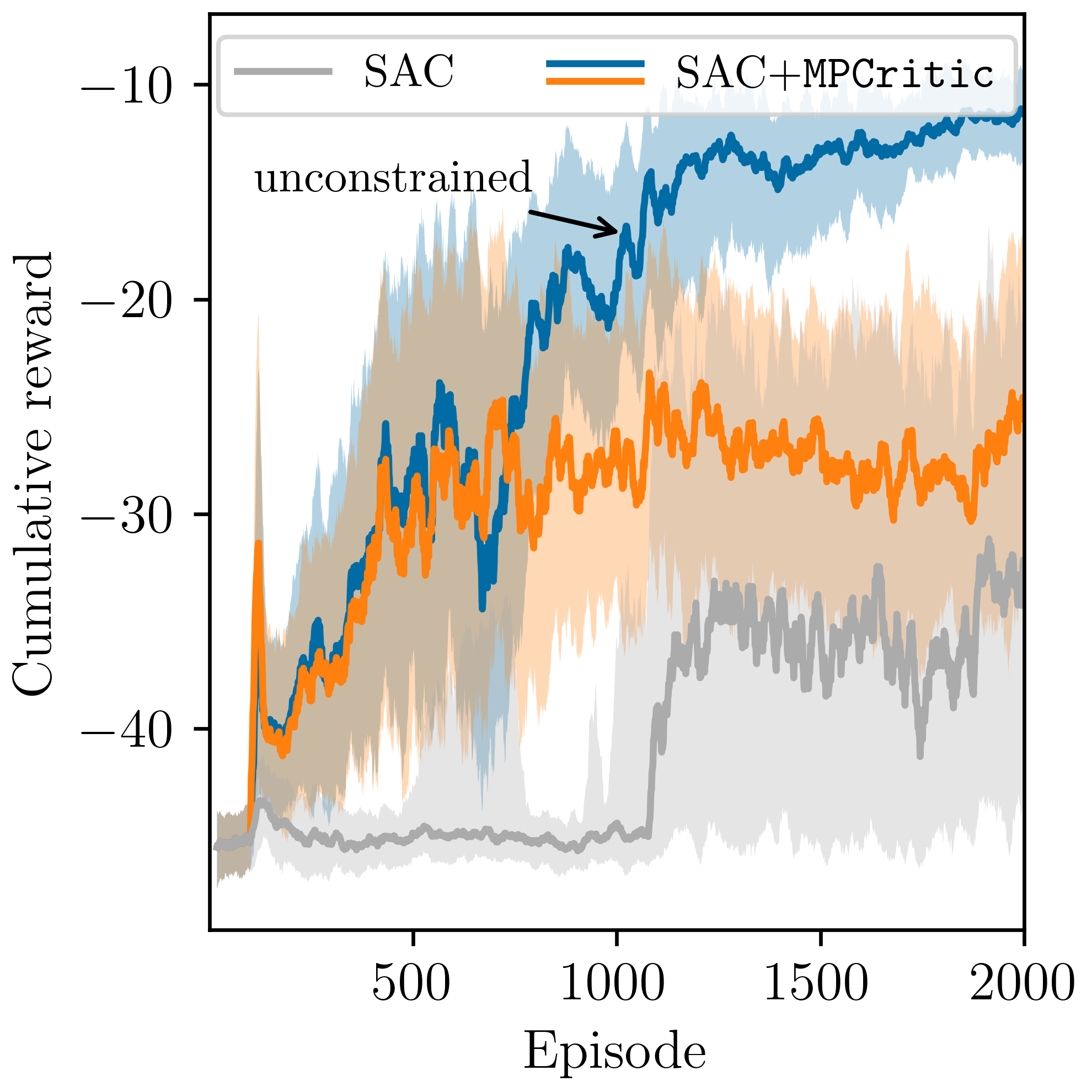}
     \caption{Cumulative reward for three \ac{SAC} agents: a default agent and two with \mpcritic to enhance the policy updates. The constrained \ac{SAC}$+\mpcritic$ includes constraints in the \mpcritic structure, not the reward function.
     }
     \label{fig:cstr_reward}
  \end{figure}

\begin{figure}[tbp]
     \centering
     \includegraphics[scale=1.0]{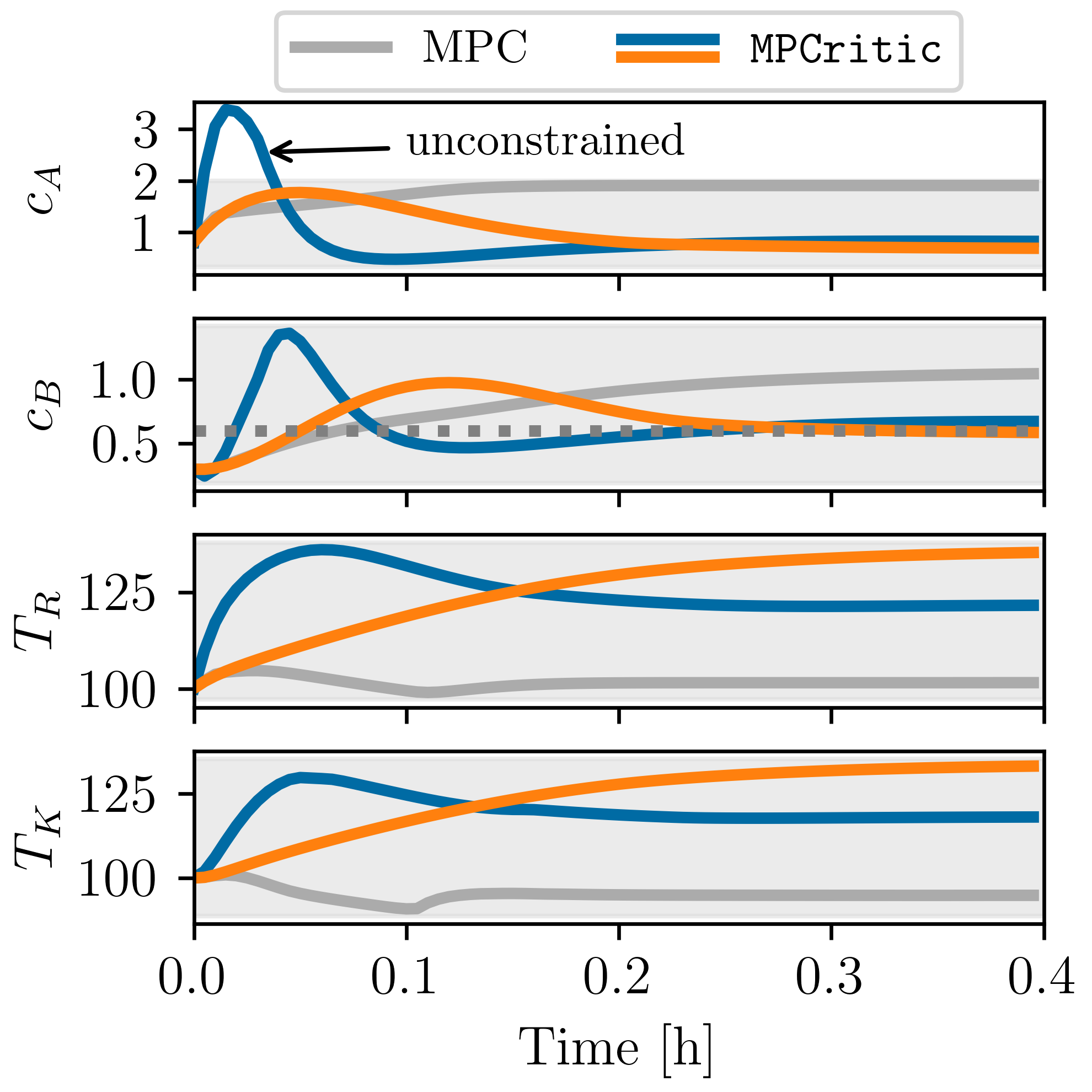}
     \caption{Trajectories from the two \mpcritic agents in \cref{fig:cstr_reward} and a robust \ac{MPC} agent (gray). The constrained \mpcritic agent is able to effectively balance the goal of the \ac{MDP} with its intrinsic constraints.
     }
     \label{fig:cstr_traj}
  \end{figure}

Because \mpcritic is a plug-and-play architecture, we can embed it directly in a default \ac{SAC} agent.
\Cref{fig:cstr_reward} shows three reward curves, each over $10$ seeds.
The vanilla \ac{SAC} agent takes over $1000$ episodes to start improving, only reaching a modest level of reward.
We note that this is not a critique of \ac{SAC}; fine-tuning its hyperparameters can indeed lead to improved results.
Rather, we stress that \mpcritic provides a useful inductive bias to jump start and enhance the learning process, indicated by the ``unconstrained'' reward curve.
Importantly, \mpcritic also incorporates constraints.
The intermediate reward curve in \cref{fig:cstr_reward} indicates that \ac{SAC}$+\mpcritic$ is able to learn a high-performing policy under the Gaussian reward, but that is fundamentally limited by the constraints in its representation.
Trajectories from both \mpcritic agents are shown in \cref{fig:cstr_traj}, along with the robust \ac{MPC} policy given by \cite{fiedler2023DompcFAIR}.
The unconstrained \mpcritic agent represents a near-optimal solution to the setpoint $c_B^{\text{goal}}$; \mpcritic with constraints also achieves its goal, taking longer, but staying within the shaded region. 
Meanwhile, the \ac{MPC} agent may achieve robust constraint satisfaction, but it contains no goal-directed feedback to improve its response when a better course of action may exist.

\section{Conclusions}

\mpcritic is an algorithmic framework capable of seamlessly utilizing advanced tools from both \ac{MPC} and \ac{RL}. 
While we have demonstrated the scalability and versatility of \mpcritic across different configurations, there are many fruitful paths for future work.
These range from formalizing theoretical properties of \mpcritic and its applications with more sophisticated \ac{MPC} agents to establishing its utility as a general inductive bias in \ac{RL} for complex environments.







\bibliographystyle{IEEEtran}
\bibliography{2025_cdc.bib}

\end{document}